\documentclass [sigconf]{acmart}

\AtBeginDocument{%
  }

\setcopyright{acmlicensed}
\copyrightyear{2026}
\acmYear{2026}
\acmDOI{XXXXXXX.XXXXXXX}
\acmConference[Conference acronym '26]{Conference name}{Month DD--DD, 2026}{City, Country}

\setcopyright{none}

\makeatletter
\renewcommand\@formatdoi[1]{\ignorespaces}
\makeatother
\renewcommand{\footnotetextcopyrightpermission}[1]{}

\usepackage{amsmath}
\usepackage{multicol,multirow}
\usepackage{booktabs}
\usepackage{xcolor}

\usepackage{makecell}
\usepackage{color,colortbl}
\definecolor{light-gray}{gray}{0.9}
\usepackage{subcaption}
\usepackage{algorithm}
\usepackage{algorithmic}
\usepackage{bm}

\begin{document}

\title{Uncertainty-Aware Token Importance Estimation in \\ Spiking Transformers}

\author{Wenxuan Liu}
\email{liuwx66@pku.edu.cn}
\affiliation{%
 \institution{School of Computer Science, \\Peking University}
 \city{Beijing}
 \country{China}
}

\author{Zecheng Hao}
\email{haozecheng@pku.edu.cn}
\affiliation{%
 \institution{School of Computer Science, \\Peking University}
 \city{Beijing}
 \country{China}
}

\author{Tong Bu}
\email{putong30@pku.edu.cn}
\affiliation{%
 \institution{School of Computer Science, \\ Peking University.\\
 Institute for Artificial Intelligence, Peking University}
 \city{Beijing}
 \country{China}
}

\author{Yuran Wang}
\email{yuranwang25@stu.pku.edu.cn}
\affiliation{%
 \institution{Peking University}
 \city{Beijing}
 \country{China}
}

\author{Zhaofei Yu}
\email{yuzf12@pku.edu.cn}
\authornote{Corresponding author}
\affiliation{%
 \institution{School of Computer Science, \\ Peking University.\\
 Institute for Artificial Intelligence, Peking University}
 \city{Beijing}
 \country{China}
}

\renewcommand{\shortauthors}{Trovato et al.}

\begin{abstract}
Spiking transformers have shown strong potential for neuromorphic vision, yet their multi-step token processing still introduces substantial redundancy and inference cost. Existing token reduction methods mainly rely on response-driven cues, such as activation magnitude, firing statistics, or feature similarity. Although effective, these criteria do not explicitly characterize token importance from the perspective of temporally evolving class evidence. In spiking transformers, token representations are progressively formed across multiple spiking steps rather than determined at a single instant, suggesting that token importance should be evaluated not only by instantaneous responses but also by temporal uncertainty patterns. Our key observation is that tokens exhibit heterogeneous uncertainty trajectories over time, and that their temporally aggregated uncertainty statistics provide an effective cue for distinguishing informative tokens from redundant ones. Motivated by this, we propose Uncert, a training-free and plug-and-play token importance estimation framework for spiking transformers. Specifically, Uncert models token-wise class evidence with a Dirichlet distribution and summarizes each token’s temporal uncertainty using its mean and fluctuation across spiking steps, yielding an uncertainty-aware importance score for inference-time token reduction. Experiments on both static and neuromorphic benchmarks show that Uncert achieves favorable accuracy-efficiency trade-offs, with the most consistent gains observed under token pruning. Further analysis reveals a clear empirical connection between temporal uncertainty patterns and token contribution, offering new insights into token dynamics in spiking transformers.
\end{abstract}

\ccsdesc[500]{Computing methodologies~Feature selection}
\ccsdesc[500]{Computing methodologies~Computer vision}
\keywords{Spiking Vision Transformer, Token Selection, Uncertainty-aware, Training-free, Energy Efficiency}

\received{20 February 2007}
\received[revised]{12 March 2009}
\received[accepted]{5 June 2009}

\maketitle

\section{Introduction}

The human brain achieves remarkable visual efficiency under strict energy constraints, partly through selective neural activation, where only a subset of neurons responds strongly to relevant stimuli while less useful responses are suppressed~\cite{vinje2000sparse,yoshida2020natural}. This selective processing mechanism suggests an appealing principle for efficient artificial perception: computational resources should be preferentially allocated to informative signals.

Inspired by such selective processing in biological vision, neuromorphic vision systems~\cite{zhou2022spikformer,zhou2024spatial} exploit event-driven sensing~\cite{maas1997networks} and spiking neural networks to achieve sparse and temporally precise perception. Building on this paradigm, spiking transformers further introduce self-attention to model long-range spatio-temporal dependencies. However, the sparsity of spikes does not automatically translate into efficient token processing: once transformed into token sequences across multiple simulation steps, spiking representations can still contain substantial redundancy, leading to nontrivial inference cost~\cite{hao2025conversion,huang2024towards,huang2025icml}.

\begin{figure}[!t]
	\centering
	\includegraphics[width = \linewidth]{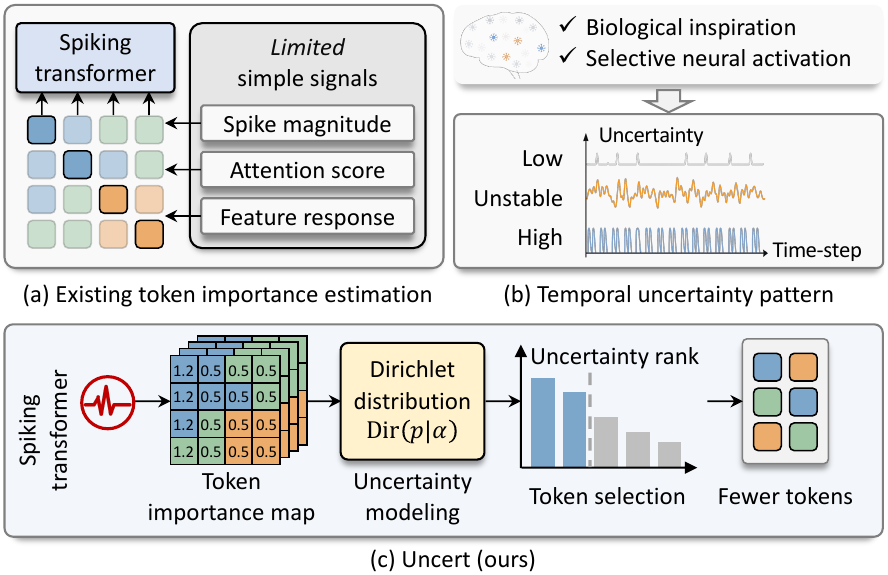}
	\caption{\textbf{Observation of Uncert.} (a) Existing token importance estimation methods in spiking transformers mainly rely on response-driven heuristics. (b) Tokens in spiking transformers exhibit distinct temporal patterns across time steps, suggesting that temporally evolving uncertainty provides informative cues for token importance estimation. (c) Uncert converts temporal uncertainty dynamics into an inference-time token importance map for efficient token reduction in spiking transformers.}
	\label{fig:1}
\end{figure}

A natural way to improve efficiency is token reduction~\cite{iclr/BolyaFDZFH23}. Existing methods typically estimate token importance from response-driven cues, such as activation magnitude, firing statistics, feature similarity, or lightweight learned predictors, as illustrated in Figure~\ref{fig:1}(a). Although effective, these criteria mainly measure token saliency from feature responses and do not explicitly characterize how token-level class evidence evolves across multiple spiking steps~\cite{cvpr/Zeng0000O022,RaoZLLZH21}. This limitation is particularly important in spiking transformers, where token representations are progressively formed through sparse neural firing rather than determined at a single instant. As a result, token importance cannot be adequately inferred from single-step responses alone, calling for a temporal perspective that captures how token-level evidence evolves throughout the spiking dynamics.

Our key observation is that token uncertainty in spiking transformers exhibits structured temporal patterns rather than uniform dynamics, as illustrated in Figure~\ref{fig:1}(b). Since token-level class evidence is progressively accumulated across spiking steps, different tokens can exhibit markedly different uncertainty trajectories over time, including stable, fluctuating, or persistently ambiguous patterns. Such heterogeneity suggests that temporal uncertainty statistics may provide a useful and previously underexplored cue for token importance estimation, beyond response strength alone. Rather than assuming a fixed monotonic relation between uncertainty and importance, we investigate whether temporally aggregated uncertainty can serve as an effective proxy for distinguishing informative tokens from redundant ones.

Motivated by this observation, we propose Uncert, a training-free and plug-and-play token importance estimation framework for spiking transformers, as illustrated in Figure~\ref{fig:1}(c). Specifically, Uncert models token-wise class evidence with a Dirichlet distribution and summarizes each token’s temporal uncertainty using its level and fluctuation across spiking steps, yielding an uncertainty-aware importance score for inference-time token reduction. The resulting score can be directly applied to token pruning, and can also be extended to token merging as an alternative reduction strategy. Experiments on both static and neuromorphic benchmarks show favorable accuracy-efficiency trade-offs, with token pruning emerging as the most effective use case.

Our contributions are summarized as follows:
\begin{itemize}
    \item We revisit token importance estimation in spiking transformers from the perspective of temporal uncertainty, and show that tokens exhibit heterogeneous uncertainty trajectories across spiking steps.
    \item We propose {Uncert}, a training-free and plug-and-play token importance estimation framework that models token-wise class evidence with a Dirichlet distribution and derives inference-time token scores from temporally aggregated uncertainty statistics.
    \item We validate Uncert on both static and neuromorphic benchmarks, demonstrating favorable accuracy-efficiency trade-offs, especially under token pruning, and providing further analysis on the relation between temporal uncertainty patterns and token contribution.
\end{itemize}

\section{Preliminaries and Related Works}
Spiking neural networks (SNNs), often considered the third generation of neural networks, have attracted growing interest due to their event-driven computation and remarkable energy efficiency. In this work, we employ the widely adopted Leaky Integrate-and-Fire (LIF) neuron~\cite{izhikevich2004model,gerstner2002spiking} model to validate the feasibility of the proposed Uncert. At each time step $t$, the membrane potential integrates incoming spikes from the previous layer while gradually decaying over time:
\begin{equation}
U_i^{l}[t] = \tau U_i^{l}[t-1] + W_i^{l} S^{l-1}[t],
\end{equation}
where $0 < \tau < 1$ denotes the membrane decay factor that models the leakage of the membrane potential. When the membrane potential exceeds the firing threshold $V_{th}$, the neuron emits a spike:
\begin{equation}
S_i^{l}[t] = H(U_i^{l}[t] - V_{th}),
\end{equation}
where $H(\cdot)$ denotes the Heaviside step function. After spike emission, the membrane potential is reset as
\begin{equation}
U_i^{l}[t] = U_i^{l}[t](1 - S_i^{l}[t]).
\end{equation}
The temporal integration and leakage process introduces variability across time steps. Early SNNs are primarily built upon convolutional architectures, which mainly focus on local feature extraction~\cite{XianMM24}. 

\textbf{Spiking Transformers.} Transformer architectures have recently demonstrated strong capability in capturing global contextual dependencies through self-attention~\cite{LiW0MXMF22,pr/LiuJZJYY25}. Inspired by the success, spiking transformers have been proposed to integrate the spike-driven computation of SNNs with the global modeling ability~\cite{yaospike}. Spikformer~\cite{zhou2022spikformer} pioneers the introduction of self-attention and enables direct Transformer training in SNNs by employing spike-based $Q$, $K$, and $V$ representations, eliminating floating-point multiplications in attention computation. 
Spikformer V2 further explores masked image modeling in spiking transformers~\cite{zhou2024spikformer}. SpikingResformer~\cite{shi2024spikingresformer} integrates dual-spike self-attention with a ResNet-style architecture, improving performance while reducing parameter complexity. QKFormer~\cite{zhou2024qkformer} adopts spike-based $Q$ and $K$ for attention computation and introduces spiking patch embeddings with deformable shortcuts. Meanwhile, Spike-driven Transformer (SDT) series~\cite{yao2023spike,yao2024spike,yao2025spike}, which introduces spike-driven attention mechanisms to enable efficient Transformer computation in SNNs. Besides, Max-Former restores high-frequency signals through frequency-enhancing operators~\cite{nips2025}.

\subsection{Token Selection in Spiking Transformers} 

Given the heavy computational cost of Transformers, several works have attempted to develop more efficient Transformer architectures for both NLP and vision tasks~\cite{iclr/BolyaFDZFH23,naacl/Lee-ThorpAEO22}. Since transformers can operate on a variable number of tokens, several recent works have attempted to prune tokens to improve the efficiency of transformers, including spiking transformers~\cite{liu2024sparsespikformer,kang2024atsnn,zhuge2024efficientspikingtransformer}. SparseSpikformer proposes a co-design framework that jointly performs token pruning and weight pruning to improve the efficiency of spiking transformers~\cite{liu2024sparsespikformer}. AT-SNN introduces an adaptive token mechanism that dynamically selects informative tokens in spiking vision transformers to reduce redundant computation~\cite{kang2024atsnn}. STATA proposes a token sparsification framework with a learnable importance predictor to remove redundant tokens during training and inference~\cite{zhuge2024efficientspikingtransformer}. TP-Former prunes tokens based on spike firing statistics, leveraging the intrinsic sparsity of spiking activations in a training-free manner~\cite{wei2026tpspikformer}. Meanwhile,~\cite{buactivity} has explored suppressing neuronal activity to improve SNN efficiency by introducing adaptive neuron AT-LIF, which dynamically adjusts firing thresholds to reduce spike rates.

Existing token pruning or merging methods in spiking transformers mainly rely on heuristic importance estimation, such as activation magnitude, firing statistics, feature similarity, or refined features. These approaches measure token importance from deterministic feature responses but overlook the intrinsic stochasticity of spiking representations. In contrast, our method estimates token importance from temporal uncertainty, providing a {uncertainty-aware criterion} for token selection.

\begin{figure}[t]
	\centering
    \includegraphics[width = 0.47\linewidth]{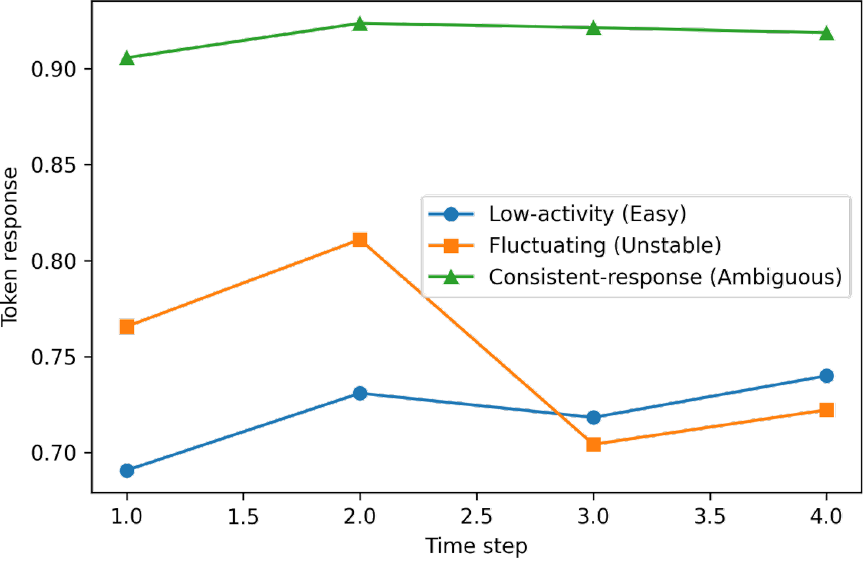}
    \label{fig:subfig_a}
  \hfill
    \includegraphics[width = 0.47\linewidth]{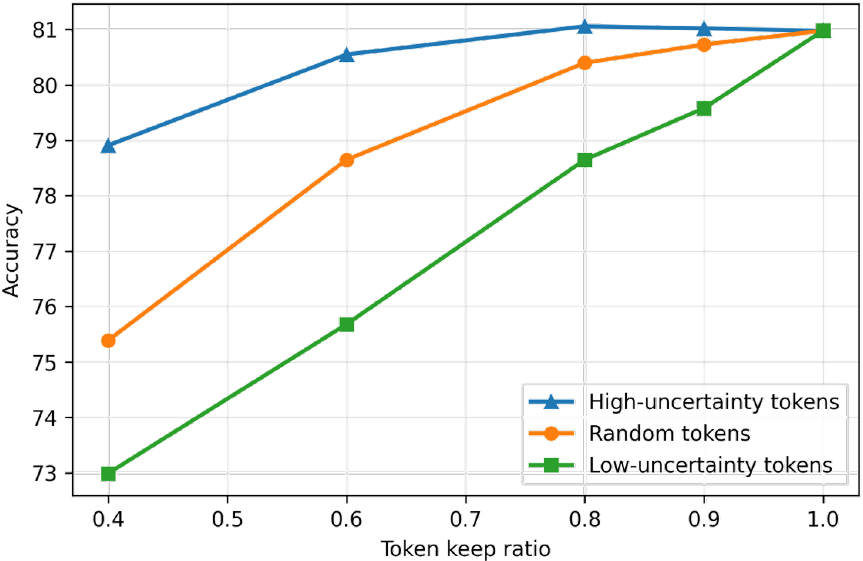} 
    \label{fig:subfig_b}
	\caption{\textbf{Motivation of Uncert.} Left: Temporal uncertainty trajectories of three representative tokens, corresponding to an easy token with consistently low uncertainty (blue), an unstable token with large fluctuations (orange), and an ambiguous token with persistently high uncertainty (green). Right: Comparison of token selection strategies, including retaining high-uncertainty tokens (blue triangles), random selection (orange circles), and retaining low-uncertainty tokens (green squares).}
	\label{fig:2}
\end{figure}

\section{Motivation}
\label{sec:motiv1}
Spiking transformers process visual information through tokens that evolve across multiple discrete time steps. Let $z_i^t$ denote the representation of token $i$ at time step $t$, where $t = 1, \dots, T$. Unlike conventional transformers that operate on static token features, spiking transformers progressively form token representations through sparse neural firing dynamics. As a result, token importance is inherently temporal rather than determined by a single-step response alone.

\subsection{Temporal Uncertainty Patterns}

To better understand the temporal variations of tokens in spiking transformers, we analyze the uncertainty trajectories of tokens across time-steps. Specifically, we compute the token-level uncertainty at each time step based on the evidence learning~\cite{shafer1976mathematical} derived from the classifier outputs. Specifically, for each token $i$ at time step $t$, we estimate an uncertainty score $U_i^t$ from the classifier responses, forming a temporal trajectory for each token: $\{U_i^t\}_{t=1}^{T}$, where $T$ denotes the number of time steps. We then visualize representative tokens selected according to their uncertainty statistics, including an easy token with consistently low uncertainty, an unstable token with large temporal fluctuations, and an ambiguous token with persistently high uncertainty. As illustrated in Figure~\ref{fig:2} (a), these tokens exhibit distinct temporal patterns. This observation indicates that token uncertainty in spiking transformers is not uniform but instead presents structured temporal behaviors. Such temporal statistics provide informative cues for identifying meaningful tokens, motivating us to leverage both the mean and variance of token uncertainty for token importance estimation. This observation highlights the central motivation of our method: tokens in spiking transformers exhibit heterogeneous temporal uncertainty patterns, suggesting that temporal uncertainty statistics may provide useful cues for token importance estimation.

\subsection{Uncertainty as a Cue for Token Importance}

To further examine whether temporal uncertainty in Sec.~\ref{sec:motiv1} is associated with token importance, we compare different token selection strategies under varying token keep ratios. Specifically, we evaluate three strategies: retaining tokens with the highest uncertainty, random token selection, and retaining tokens with the lowest uncertainty. Rather than assuming a fixed deterministic relation, we empirically investigate whether uncertainty can serve as an effective cue for identifying informative tokens. As shown in Figure~\ref{fig:2}(b), retaining high-uncertainty tokens consistently yields higher accuracy than random selection and low-uncertainty token selection, especially at lower keep ratios. This result suggests that token uncertainty is not merely a measure of instability, but may also reflect the presence of informative and discriminative content. Empirically, tokens with higher uncertainty are more likely to correspond to regions where the model faces greater ambiguity yet receives task-relevant evidence, whereas low-uncertainty tokens are more often associated with trivial or redundant content (\textit{e.g.,} background regions). Therefore, uncertainty provides a useful signal for estimating token importance in spiking transformers.

\begin{figure*}[h]
    \centering
	\includegraphics[width = \linewidth]{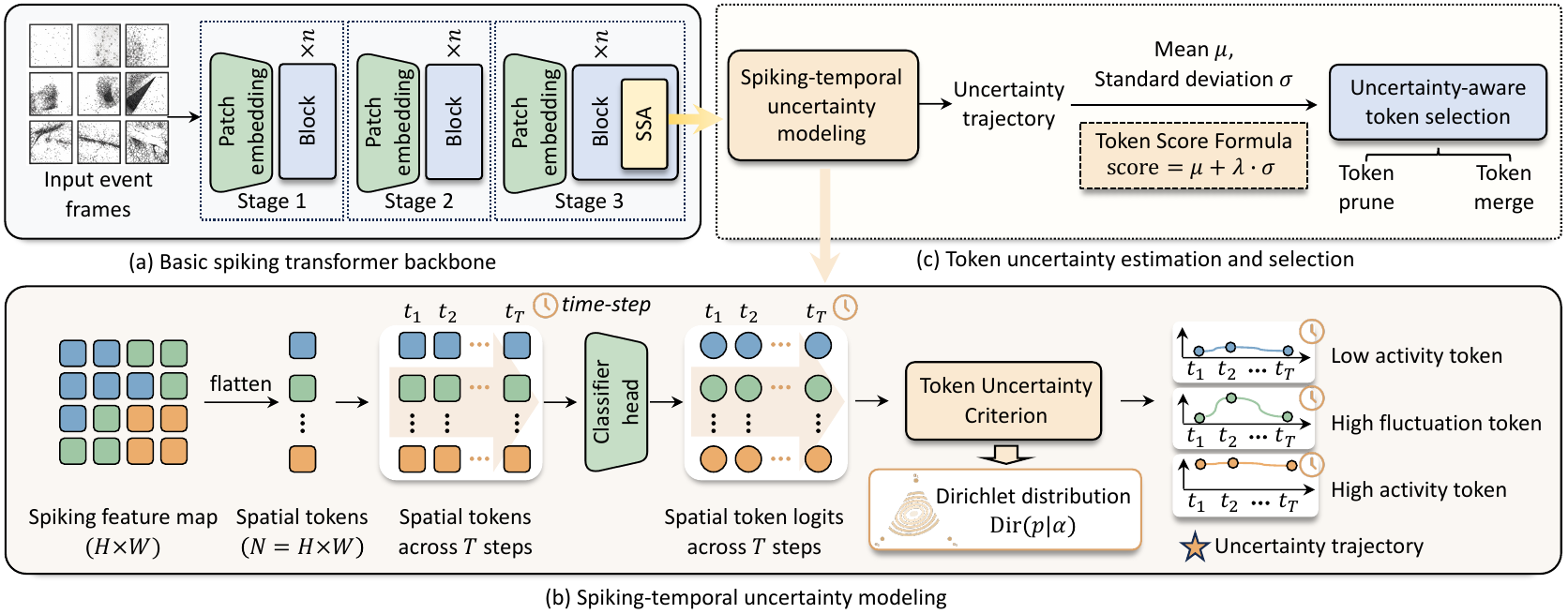}
	\caption{The overall framework of the proposed Uncert, including the basic spiking transformer backbone (top) and the spiking token uncertainty estimation (bottom). (a) shows the spiking transformer backbone, instantiated with QKFormer~\cite{zhou2024qkformer} or MaxFormer~\cite{nips2025}. (b) presents the proposed spiking-temporal uncertainty modeling. (c) shows uncertainty-aware token selection for pruning and merging.}
	\label{fig:method}
\end{figure*}

\section{Method}
In this section, we introduce our Uncertainty-aware token importance estimation (Uncert) for efficient spiking
transformers, as illustrated in Figure~\ref{fig:method}. Our framework consists of four components: spiking token hierarchy, token uncertainty criterion, spiking-temporal uncertainty modeling, and token selection.

\subsection{Backbone and Spiking Token Hierarchy}
We adopt a QKFormer-style spiking transformer as the backbone~\cite{zhou2024qkformer}. Given an input sequence $\mathcal{I}\in\mathbb{R}^{T_0 \times B \times n \times H \times W}$, where $T_0$ denotes the number of input time steps, $B$ is the batch size, and $n$ is the input channel dimension, we consider two types of inputs. For conventional static RGB images, the temporal dimension collapses to a single step ($T_0=1$) with three channels ($n=3$), and the input frame is repeated over $T$ simulation steps to drive the spiking dynamics. For neuromorphic inputs such as event-based frame data, the input naturally contains multiple temporal steps ($T_0=T$) with two polarity channels ($n=2$).

The backbone consists of three hierarchical stages. We denote the stage-wise spiking feature map as
\begin{equation}
X^{(s)} \in \mathbb{R}^{T \times B \times C_s \times H_s \times W_s}, \quad s=1,2,3,
\end{equation}
where $C_s$, $H_s$, and $W_s$ represent the channel dimension and spatial resolution at stage $s$, respectively. 
The input sequence is first projected into spiking feature maps through an initial patch embedding module. 
After flattening the spatial dimensions, the stage-wise feature maps can be equivalently represented in token form as
\begin{equation}
Z^{(s)} \in \mathbb{R}^{T \times B \times N_s \times D_s},
\end{equation}
where $N_s = H_sW_s$ denotes the number of tokens and $D_s=C_s$ is the token dimension. Specifically, Stage 1 applies an initial spiking patch embedding module followed by a Token-QK Transformer block. Stage 2 further performs spatial downsampling and channel expansion through a patch embedding stage, followed by another Token-QK Transformer block. Stage 3 applies an additional patch embedding stage and then processes the features using multiple spiking transformer blocks with spike-driven self-attention. 

As the network depth increases, the number of tokens gradually decreases while the channel dimension increases, forming a hierarchical spatio-temporal spiking token representation. Based on these temporally evolving token features, we estimate token uncertainty across time steps to infer token importance.

\subsection{Token Uncertainty Criterion}

Let $z_i^t \in \mathbb{R}^{D}$ denote the representation of token $i$ at time step $t$, where $i=1,\dots,N$ indexes the spatial tokens and $t=1,\dots,T$ denotes the simulation time steps. To characterize token-level class evidence, we estimate temporal uncertainty from classifier responses using a Dirichlet distribution.

Specifically, given token representation $z_i^t$, the classifier head first produces class logits
\begin{equation}
l_{i,c}^t = g_c(z_i^t),
\end{equation}
which are converted into non-negative evidence values using the Softplus activation,
\begin{equation}
e_{i,c}^t = \mathrm{Softplus}(l_{i,c}^t) = \log(1+\exp(l_{i,c}^t)),
\end{equation}
where $\mathrm{Softplus}$ is adopted to ensure that the evidence values are non-negative. The Dirichlet parameters are then defined as
\begin{equation}
\alpha_{i,c}^t = e_{i,c}^t + 1.
\end{equation}
Let
\begin{equation}
S_i^t = \sum_{c=1}^{C} \alpha_{i,c}^t
\end{equation}
denote the total evidence. Following evidential uncertainty modeling, the uncertainty of token $i$ at time step $t$ is defined as
\begin{equation}
U_i^t = \frac{C}{S_i^t}.
\end{equation}
This formulation yields a temporal uncertainty trajectory $\{U_i^t\}_{t=1}^{T}$ for each token, reflecting how the concentration of its class evidence evolves across spiking time steps.

\begin{table*}[t]
\centering
\caption{Comparison with state-of-the-art methods on both static and neuromorphic datasets, CIFAR-10, CIFAR-100, DVS-Gesture, and DVS-CIFAR-10. We report the number of parameters (Param. (M)), the number of time steps ($T$), and the top-1 accuracy (Acc@1 (\%)).}
\label{tab:sota}
\definecolor{myorange}{RGB}{220,120,20}
\setlength{\tabcolsep}{3pt}
\begin{tabular}{l|ccc|ccc|ccc|ccc}
\toprule
\multirow{2}{*}{\textbf{Method}} 
& \multicolumn{3}{c|}{\textbf{CIFAR-10}} 
& \multicolumn{3}{c|}{\textbf{CIFAR-100}} 
& \multicolumn{3}{c|}{\textbf{DVS-Gesture}} 
& \multicolumn{3}{c}{\textbf{DVS-CIFAR-10}} \\
\cmidrule(lr){2-4} \cmidrule(lr){5-7} \cmidrule(lr){8-10} \cmidrule(lr){11-13}
& Param. (M) & $T$ & Acc@1 (\%)
& Param. (M) & $T$ & Acc@1 (\%) 
& Param. (M) & $T$ & Acc@1 (\%)
& Param. (M) & $T$ & Acc@1 (\%) \\
\midrule
Spikformer~\cite{zhou2022spikformer} 
& 9.32 & 4 & 95.51 
& 9.32 & 4 & 78.21 
& 2.57 & 16 & 98.30 
& 2.57 & 16 & 80.90 \\

S-Transformer~\cite{yao2023spike} 
& 10.28 & 4 & 95.60 
& 10.28 & 4 & 78.40 
& 2.57 & 16 & 99.30 
& 2.57 & 16 & 80.00 \\

SWformer~\cite{fang2024spikingeccv} 
& 7.51 & 4 & 96.10 
& 7.51 & 4 & 79.30 
& - & - & - 
& 2.05 & 16 & 83.90 \\

STAtten~\cite{lee2025spiking} 
& - & 4 & 96.03 
& - & 4 & 80.20 
& - & - & - 
& - & 16 & 83.90 \\

\midrule

QKFormer~\cite{zhou2024qkformer} 
& 6.74 & 4 & 96.31 
& 6.74 & 4 & 80.98 
& 1.99 & 16 & 98.61
& 1.50 & 10 & 82.30 \\

\cellcolor{myorange!8}Uncert (Ours)
&\cellcolor{myorange!8}6.74 & \cellcolor{myorange!8}4 &  \cellcolor{myorange!8}96.30
& \cellcolor{myorange!8}6.74 & \cellcolor{myorange!8}4 & \cellcolor{myorange!8}81.06
& \cellcolor{myorange!8}1.99 & \cellcolor{myorange!8}16 & \cellcolor{myorange!8}96.88
& \cellcolor{myorange!8}1.50 & \cellcolor{myorange!8}10 &  \cellcolor{myorange!8}82.50 \\
\bottomrule
\end{tabular}
\end{table*}

\subsection{Spiking-temporal Uncertainty Modeling}
We summarize each token’s temporal uncertainty trajectory using its mean and standard deviation across spiking steps.

Unlike conventional feature-based temporal metrics, we characterize token dynamics through uncertainty trajectories. 
Specifically, given the uncertainty sequence $\{U_i^t\}_{t=1}^{T}$ for token $i$, we summarize its temporal behavior using the mean and standard deviation:
\begin{equation}
\mu_i = \frac{1}{T}\sum_{t=1}^{T} U_i^t, \sigma_i = \sqrt{\frac{1}{T}\sum_{t=1}^{T}(U_i^t-\mu_i)^2 } .
\end{equation}
Here, $\mu_i$ measures the overall ambiguity of the token, while $\sigma_i$ captures the temporal fluctuation of its uncertainty, indicating whether the token responses are stable or highly variable during the spiking dynamics.
Empirically, tokens with higher uncertainty levels or stronger temporal fluctuations are more likely to correspond to informative regions where the model still needs to resolve ambiguity. These temporal statistics provide complementary cues for identifying informative tokens.

{Based on temporal uncertainty statistics, we estimate token importance for downstream recognition. Intuitively, tokens with relatively higher uncertainty or stronger temporal fluctuations are more likely to correspond to semantically informative regions, where the model needs to resolve ambiguity and capture discriminative cues. We therefore define the token importance score as:}
\begin{equation}
S_i = \mu_i + \lambda \sigma_i ,
\end{equation}
where $\lambda$ is a balancing coefficient controlling the contribution of temporal fluctuation. In our implementation, $\lambda$ is empirically set to $0.9$. Tokens with larger scores are treated as more informative for inference-time token reduction. These scores are subsequently used to rank tokens and guide the token selection process.

\subsection{Token Selection}
\textbf{Token Prune.} Based on the uncertainty-aware token importance scores $\{S_i\}_{i=1}^{N}$, we perform token reduction inside the spiking self-attention (SSA) computation to reduce redundant tokens while preserving informative ones. Specifically, we select a subset of informative tokens $\mathcal{I_\text{keep}}$ according to the top-$K$ importance scores:
\begin{equation}
\mathcal{I_\text{keep}} = \operatorname{TopK}(S, N_\text{keep}), \quad {N_\text{keep}}=\lfloor rN \rfloor ,
\end{equation}
where $r$ denotes the token keep ratio and $N$ is the total number of tokens. During the SSA computation, only the selected tokens $X_{t,\mathcal{I}_\text{keep}}^{\ell-1}$ are used to compute the query, key, and value representations:
\begin{equation}
\bm{Q} = f_q(X_{t,\mathcal{I}_\text{keep}}^{\ell-1}), \quad
\bm{K} = f_k(X_{t,\mathcal{I}_\text{keep}}^{\ell-1}), \quad
\bm{V} = f_v(X_{t,\mathcal{I}_\text{keep}}^{\ell-1}),
\end{equation}
\begin{equation}
X_t^{\ell} = (\bm{Q}\bm{K}^\top)\bm{V} .
\end{equation}
This design avoids unnecessary attention computation for low-importance tokens while preserving informative tokens
for feature aggregation. Note that after the attention computation, the updated token features are scattered back to their original spatial locations to restore the dense feature layout.

\noindent\textbf{Token Merge.} In addition to token pruning, we also explore an alternative token merging strategy, where low-importance tokens are
aggregated into nearby informative tokens rather than being discarded. The merged token representation is computed as:
\begin{equation}
z_i' = \sum_{j \in \mathcal{N}(i)} w_{ij} z_j ,
\end{equation}
where $\mathcal{N}(i)$ denotes the set of tokens merged into token $i$, and $w_{ij}$ represents normalized similarity weights.

Both pruning and merging strategies are guided by the same uncertainty-aware token importance scores. In our experiments, token pruning is mainly adopted for efficiency, while token merging serves as an alternative strategy for better information preservation.

\section{Experiment}
To systematically assess the proposed method, we investigate the following four research questions from the perspectives of effectiveness, mechanism analysis, component contribution, and cross-architecture generalization.

\textit{\textbf{RQ1:}} Does temporal uncertainty provide an effective criterion for token importance estimation compared with existing methods?

\textit{\textbf{RQ2:}} How do different token selection mechanisms (e.g., random, pruning, and merging strategies) affect performance?

\textit{\textbf{RQ3:}} How do different temporal uncertainty components (e.g., temporal mean and fluctuation) contribute to the final performance?

\textit{\textbf{RQ4:}} How well does the proposed method generalize across different backbone architectures?

\subsection{Setup}
\noindent\textbf{Models and Datasets.} We evaluate our approach on two static image datasets, {CIFAR-10}~\cite{krizhevsky2009learning} and {CIFAR-100}~\cite{krizhevsky2009learning}, and two dynamic vision sensor datasets, {DVS-CIFAR-10}~\cite{li2017cifar10} and {DVS-Gesture}~\cite{amir2017}, to assess its generalization across both frame-based and event-based data. We instantiate \textit{Uncert} on two representative spiking transformer backbones, MaxFormer~\cite{nips2025} and QKFormer~\cite{zhou2024qkformer}.

\noindent\textbf{Experimental Details}
We conduct all the experiments on one NVIDIA RTX 4090 GPU with 24 GB of memory, providing sufficient computational capacity for handling high-dimensional data. For CIFAR-10 and CIFAR-100, the input resolution is set to $32 \times 32$. For DVS-based datasets, we use frame representations constructed from event streams; the spatial resolution is $128 \times 128$ for DVS-CIFAR-10. Uncert is integrated into Spiking Self-Attention stage 3.1 of QKFormer for both~{CIFAR-10} and ~{CIFAR-100} datasets.
\begin{table*}[t]
\centering
\caption{Comparison with different token selection strategy on CIFAR-10, CIFAR-100, and DVS-CIFAR-10. $N_{avg}$ denotes the average token retention ratio. 
Best or notable results are highlighted in color for visual comparison. $S$ stands for the training stage. Random is used to prune in stage3.1 here.}

\setlength{\tabcolsep}{9pt}

\definecolor{myblue}{RGB}{40,90,200}
\definecolor{mygreen}{RGB}{0,150,70}
\definecolor{myorange}{RGB}{220,120,20}
\definecolor{mypurple}{RGB}{140,40,220}
\definecolor{lightgreen}{RGB}{223,242,218}
\definecolor{lightorange}{RGB}{245,232,214}
\definecolor{lightpurple}{RGB}{230,228,245}

\begin{tabular}{l|c|ccl|ccl|ccl}
\toprule
\multirow{2}{*}{\textbf{Method}} & \multirow{2}{*}{\textbf{S}} 
& \multicolumn{3}{c|}{\textbf{CIFAR-10}} 
& \multicolumn{3}{c|}{\textbf{CIFAR-100}}
& \multicolumn{3}{c}{\textbf{DVS-CIFAR-10}} \\
\cmidrule(lr){3-5} \cmidrule(lr){6-8} \cmidrule(lr){9-11}
& 
& \textbf{T} & $N_{avg}$ & \textbf{Acc@1 (\%)} 
& \textbf{T} & $N_{avg}$ & \textbf{Acc@1 (\%)}
& \textbf{T} & \textbf{$N_{avg}$} & \textbf{Acc@1 (\%)} \\
\midrule

QKFormer~\cite{zhou2024qkformer} 
& \textbf{\checkmark}  
& 4 & $\times 1$    & 96.31 
& 4 & $\times 1$    & 80.98
& 10 & $\times 1$   & 82.30 \\
\midrule

\multirow{4}{*}{Random} & \textbf{\texttimes} 
& 4 & \cellcolor{mygreen!8}$\approx 0.80$ & \cellcolor{mygreen!8}96.15{\color{mygreen}\scriptsize{-0.16}}
& 4 & \cellcolor{mygreen!8}$\approx 0.80$ & \cellcolor{mygreen!8}80.40{\color{mygreen}\scriptsize{-0.58}}
& 10 & \cellcolor{mygreen!6}$\approx 0.90$ & \cellcolor{mygreen!6}81.10{\color{mygreen}\scriptsize{-1.20}}\\
& \textbf{\texttimes} 
& 4 & \cellcolor{mygreen!12}$\approx 0.60$ & \cellcolor{mygreen!12}95.87
& 4 & \cellcolor{mygreen!12}$\approx 0.60$ & \cellcolor{mygreen!12}78.65
& 10 & \cellcolor{mygreen!8}$\approx 0.80$ & \cellcolor{mygreen!8}80.50 \\
 & \textbf{\texttimes} 
& 4 & \cellcolor{mygreen!16}$\approx 0.40$ & \cellcolor{mygreen!16}{95.27}
& 4 & \cellcolor{mygreen!16}$\approx 0.40$ & \cellcolor{mygreen!16}75.39
& 10 & \cellcolor{mygreen!12}$\approx 0.60$ & \cellcolor{mygreen!12}75.50 \\
& \textbf{\texttimes} 
& 4 & \cellcolor{mygreen!20}$\approx 0.20$ & \cellcolor{mygreen!20}{94.61}
& 4 & \cellcolor{mygreen!20}$\approx 0.20$ & \cellcolor{mygreen!20}68.18
& 10 & \cellcolor{mygreen!16}$\approx 0.40$ & \cellcolor{mygreen!16}60.70 \\

\midrule

\multirow{4}{*}{Uncert (Merge)} & \textbf{\texttimes} 
& 4 & \cellcolor{mygreen!8}$\approx 0.80$ & \cellcolor{mygreen!8}95.80{\color{mygreen}\scriptsize{-0.51}}
& 4 & \cellcolor{mygreen!8}$\approx 0.80$ & \cellcolor{mygreen!8}77.97{\color{mygreen}\scriptsize{-3.01}}
& 10 & \cellcolor{mygreen!6}$\approx 0.90$ & \cellcolor{mygreen!6}81.10{\color{mygreen}\scriptsize{-1.20}} \\

& \textbf{\texttimes} 
& 4 & \cellcolor{mygreen!12}$\approx 0.60$ & \cellcolor{mygreen!12}{95.24}
& 4 & \cellcolor{mygreen!12}$\approx 0.60$ & \cellcolor{mygreen!12}74.36
& 10 & \cellcolor{mygreen!8}$\approx 0.80$ & \cellcolor{mygreen!8}78.90 \\

 & \textbf{\texttimes} 
& 4 & \cellcolor{mygreen!16}$\approx 0.40$ & \cellcolor{mygreen!16}{94.75}
& 4 & \cellcolor{mygreen!16}$\approx 0.40$ & \cellcolor{mygreen!16}70.41
& 10 & \cellcolor{mygreen!12}$\approx 0.60$ & \cellcolor{mygreen!12}75.10 \\

& \textbf{\texttimes} 
& 4 & \cellcolor{mygreen!20}$\approx 0.20$ & \cellcolor{mygreen!20}{94.01}
& 4 & \cellcolor{mygreen!20}$\approx 0.20$ & \cellcolor{mygreen!20}66.93
& 10 & \cellcolor{mygreen!16}$\approx 0.40$ & \cellcolor{mygreen!16}69.40 \\

\midrule

\multirow{4}{*}{Uncert (Prune)} & \textbf{\texttimes} 
& 4 & \cellcolor{mygreen!8}$\approx 0.80$ & \cellcolor{mygreen!8}96.21{\color{mygreen}\scriptsize{-0.10}}
& 4 & \cellcolor{mygreen!8}$\approx 0.80$ & \cellcolor{mygreen!8}81.06{\color{myorange}\scriptsize{+0.08}}
& 10 & \cellcolor{mygreen!6}$\approx 0.90$ & \cellcolor{mygreen!6}82.50{\color{myorange}\scriptsize{+0.20}} \\

& \textbf{\texttimes} 
& 4 & \cellcolor{mygreen!12}$\approx 0.60$ & \cellcolor{mygreen!12}96.10
& 4 & \cellcolor{mygreen!12}$\approx 0.60$ & \cellcolor{mygreen!12}80.55
& 10 & \cellcolor{mygreen!8}$\approx 0.80$ & \cellcolor{mygreen!8}81.20\\

 & \textbf{\texttimes} 
& 4 & \cellcolor{mygreen!16}$\approx 0.40$ & \cellcolor{mygreen!16}{95.91} 
& 4 & \cellcolor{mygreen!16}$\approx 0.40$ & \cellcolor{mygreen!16}78.91
& 10 & \cellcolor{mygreen!12}$\approx 0.60$ & \cellcolor{mygreen!12}74.10\\

 & \textbf{\texttimes} 
& 4 & \cellcolor{mygreen!20}$\approx 0.20$ & \cellcolor{mygreen!20}{94.97}
& 4 & \cellcolor{mygreen!20}$\approx 0.20$ & \cellcolor{mygreen!20}72.13
& 10 & \cellcolor{mygreen!16}$\approx 0.40$ & \cellcolor{mygreen!16}55.40 \\

\bottomrule
\end{tabular}
\label{tab:main_compare}
\end{table*}
\subsection{Comparison with SOTAs (RQ1)}
Table~\ref{tab:sota} compares Uncert with state-of-the-art models on CIFAR-10, CIFAR-100 and DVS-CIFAR-10. Overall, Uncert serves as a lightweight, training-free, and temporally aware token pruning strategy that achieves effective token reduction across different data modalities. Notably, by retaining only about 80\%--90\% of tokens, Uncert preserves classification performance with almost no degradation, and even yields slight improvements on CIFAR-100 and DVS-CIFAR-10. The stronger result on CIFAR-100 highlights the advantage of our criterion in more challenging classification settings, where fine-grained discrimination demands more effective token selection. Our temporal evidential uncertainty better identifies consistently informative tokens and suppresses redundant or noisy activations. The improvement on DVS-CIFAR-10 further shows that this criterion generalizes well to event-based data, where temporally aggregated evidence is important to dynamic scene.

\begin{table}[t]
\centering
\caption{Comparison of different temporal token prune scoring strategies at two keep ratios on {CIFAR-100}.}
\label{tab:temporal_short}
\small
\definecolor{mygreen}{RGB}{0,150,70}
\definecolor{myorange}{RGB}{220,120,20}
\begin{tabular}{l|c|cll}
\toprule
\textbf{Method} & \textbf{Keep ratio} & \textbf{T} & \textbf{Acc@1 (\%)} & \textbf{Acc@5 (\%)} \\
\midrule
QKFormer~\cite{zhou2024qkformer} & 1.0 & 4 & 80.98 & 95.13 \\
\midrule
\multirow{2}{*}{Uncertainty Only}
& \cellcolor{mygreen!8}0.8 & \cellcolor{mygreen!8}4 & \cellcolor{mygreen!8}80.89 & \cellcolor{mygreen!8}95.38 \\
& \cellcolor{mygreen!12}0.6 & \cellcolor{mygreen!12}4 & \cellcolor{mygreen!12}80.75 & \cellcolor{mygreen!12}95.29 \\
\midrule
\multirow{2}{*}{Mean Only}
& \cellcolor{mygreen!8}0.8 & \cellcolor{mygreen!8}4 & \cellcolor{mygreen!8}80.96 & \cellcolor{mygreen!8}95.32 \\
& \cellcolor{mygreen!12}0.6 & \cellcolor{mygreen!12}4 & \cellcolor{mygreen!12}80.53 & \cellcolor{mygreen!12}95.27 \\
\midrule
\multirow{2}{*}{Std Only}
& \cellcolor{mygreen!8}0.8 & \cellcolor{mygreen!8}4 & \cellcolor{mygreen!8}79.33 & \cellcolor{mygreen!8}95.00 \\
& \cellcolor{mygreen!12}0.6 & \cellcolor{mygreen!12}4 & \cellcolor{mygreen!12}76.20 & \cellcolor{mygreen!12}94.11\\
\midrule
\multirow{2}{*}{Uncert}
& \cellcolor{mygreen!8}0.8 & \cellcolor{mygreen!8}4 & \cellcolor{mygreen!8}\textbf{81.06}{\color{myorange}\scriptsize{+0.08}} & \cellcolor{mygreen!8}\textbf{95.45}{\color{myorange}\scriptsize{+0.32}} \\
& \cellcolor{mygreen!12}0.6 & \cellcolor{mygreen!12}4 & \cellcolor{mygreen!12}\textbf{80.55} & \cellcolor{mygreen!12}\textbf{95.23} \\
\bottomrule
\end{tabular}
\end{table}

\subsection{Ablation of Token Selection Strategy (RQ2)}

Table~\ref{tab:main_compare} presents a strategy ablation on token reduction mechanisms, comparing random selection, uncertainty-guided merging, and uncertainty-guided pruning across CIFAR-10, CIFAR-100, and DVS-CIFAR-10. Overall, uncertainty-guided pruning yields the most favorable trade-off between token retention and accuracy. On CIFAR-100, for example, at an average retention ratio of approximately 0.80, random selection achieves 80.40\%, whereas Uncert (Prune) reaches 81.06\%, outperforming random selection by 0.66 points and even surpassing the baseline accuracy of 80.98\%. At a more aggressive retention ratio of approximately 0.60, Uncert (Prune) still preserves 80.55\%, which is 1.90 points higher than random selection. In contrast, Uncert (Merge) drops to 77.97\% and 74.36\% under the same two settings, suggesting that the proposed uncertainty estimate is more suitable for identifying removable tokens than for aggregating them. A similar trend is observed on DVS-CIFAR-10, where Uncert (Prune) achieves 82.50\% at approximately 0.90 retention, improving over both the baseline (82.30\%) and random selection (81.10\%). These results demonstrate that temporal uncertainty provides an effective criterion for token selection, and that pruning is a more effective use of this criterion than merging.

\begin{figure}[t]
	\centering
\subfloat[{{CIFAR-100}.}]{
    \includegraphics[width = 0.47\linewidth]{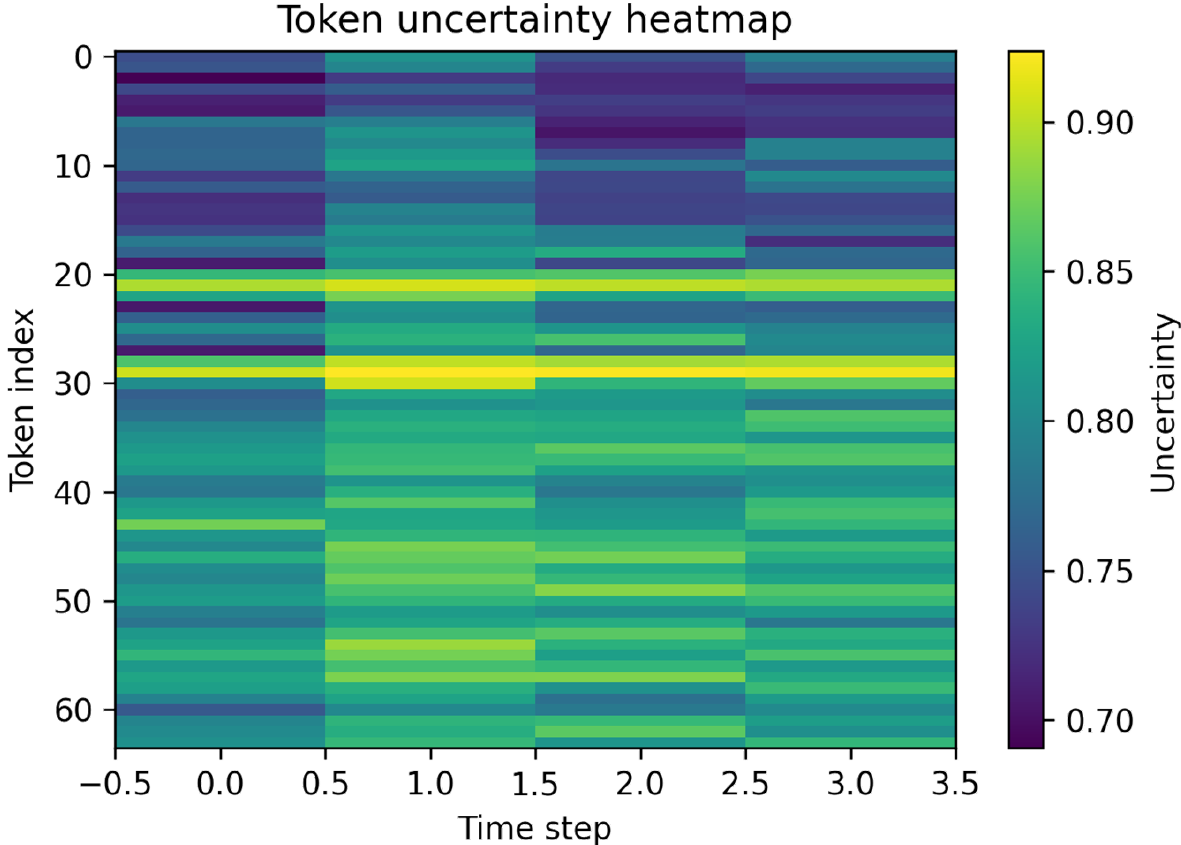}
    \label{fig:sub_a}
  } \hfill
  \subfloat[{{CIFAR-10}.}]{
    \includegraphics[width = 0.47\linewidth]{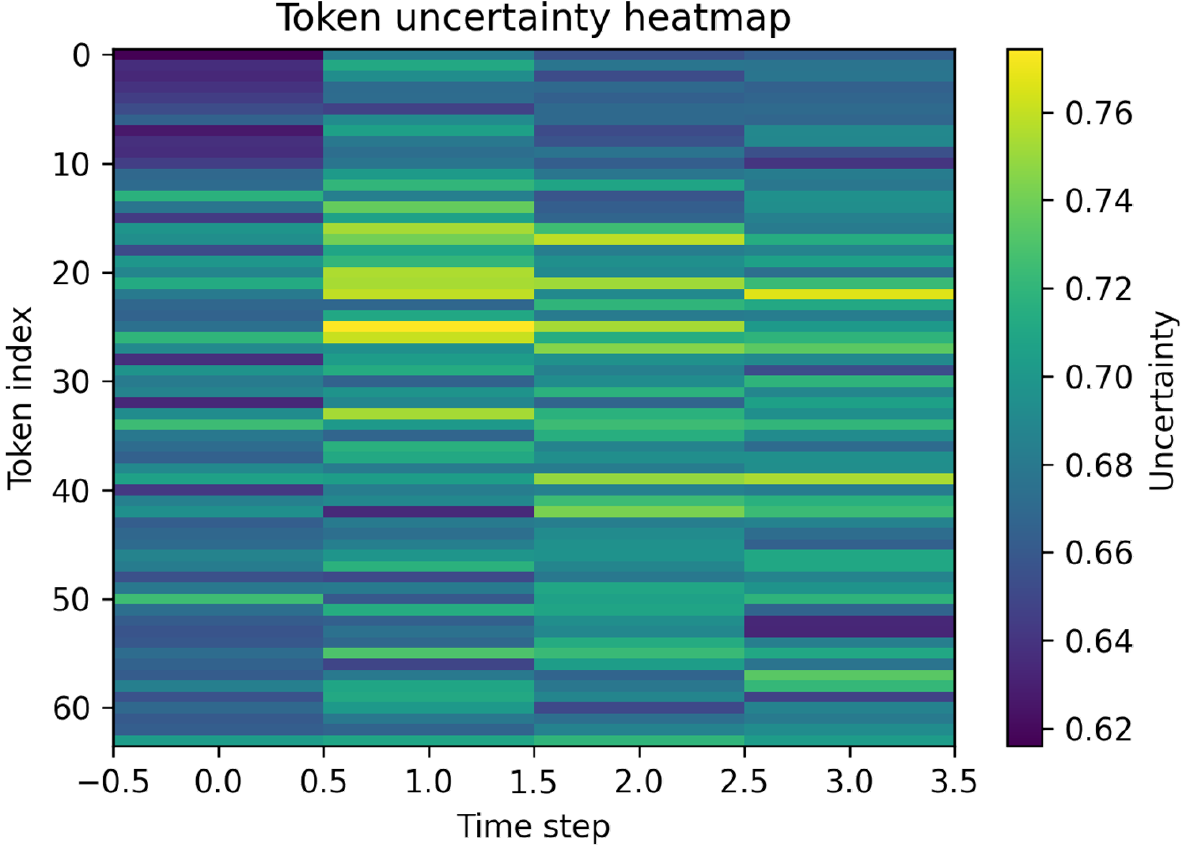} 
    \label{fig:sub_b}  } 
	\caption{\textbf{Visualization of temporal token uncertainty. The uncertainty distribution is highly non-uniform over both token index and time step, indicating that token responses exhibit heterogeneous temporal uncertainty patterns.}}
	\label{uncertainty_heatmap}
\end{figure}

\begin{figure}[t]
	\centering
\subfloat[{\textrm{Keeping ratio of 0.9.}}]{
    \includegraphics[width = 0.47\linewidth]{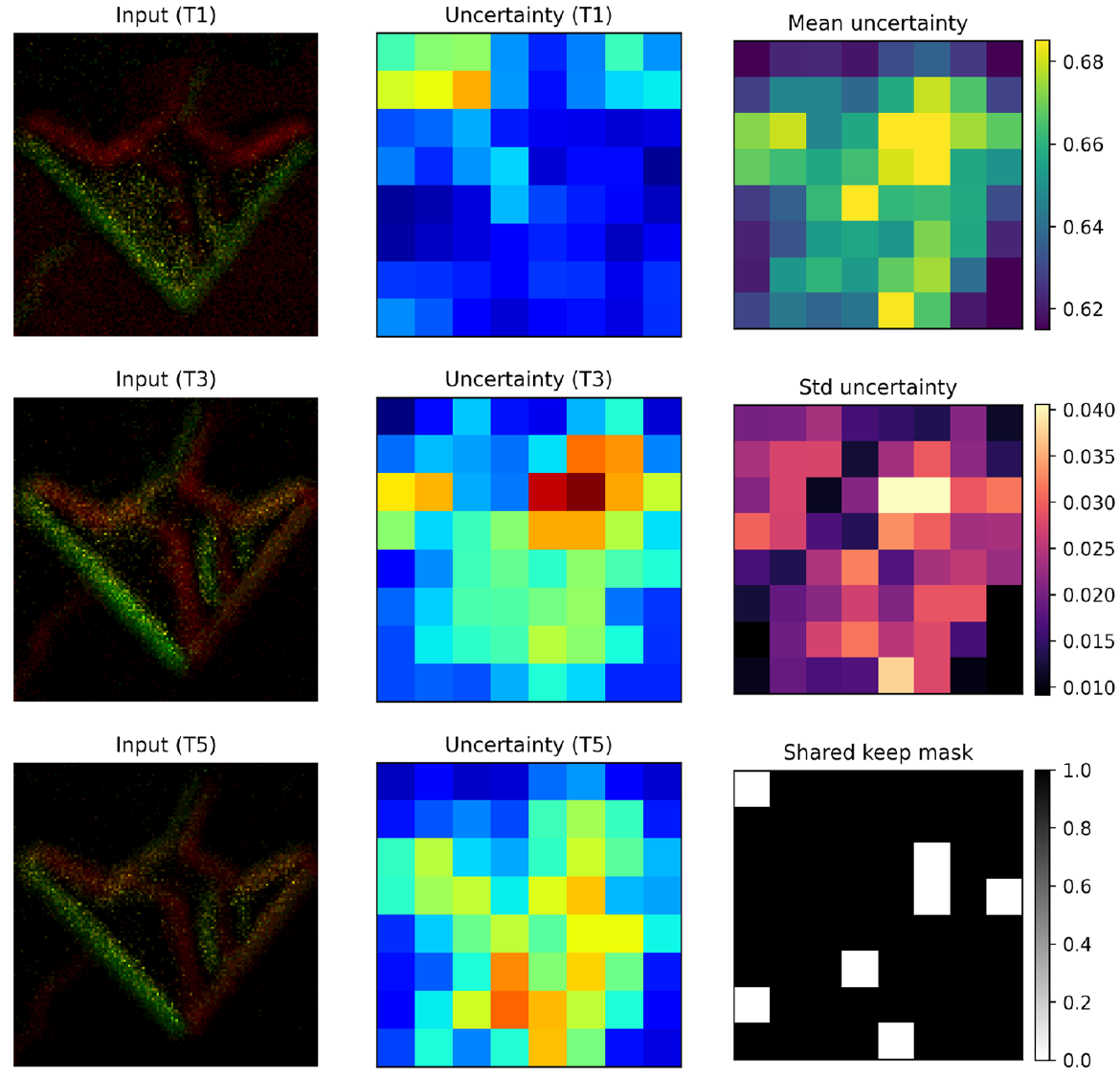}
    \label{fig:sub_0.9}
  } \hfill
  \subfloat[{Keeping ratio of 0.6.}]{
    \includegraphics[width = 0.47\linewidth]{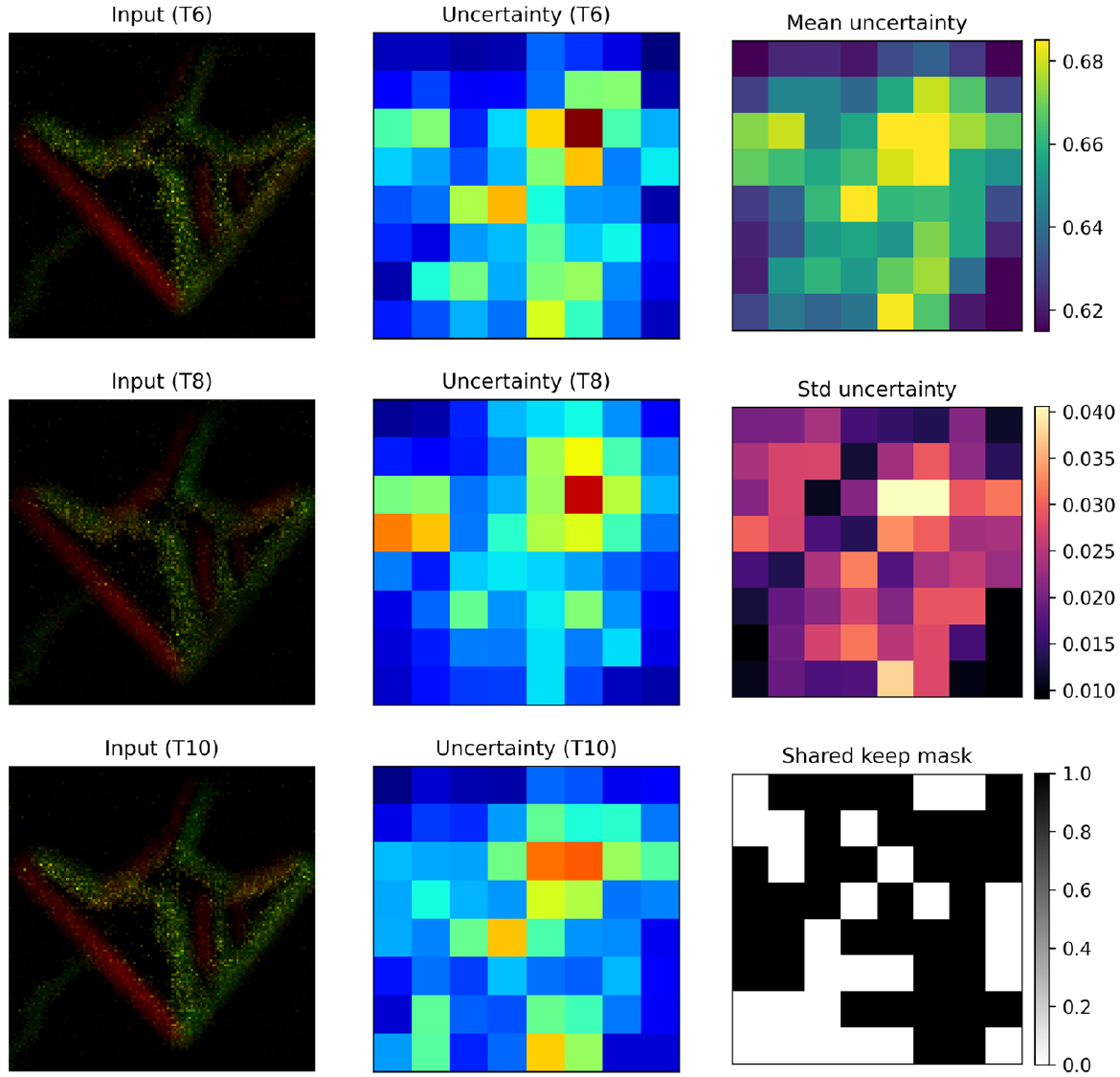} 
    \label{fig:sub_0.6}  } 
    \caption{\textbf{Pruning visualization of Uncert (QKFormer) on DVS-CIFAR-10.} Left: input frames at selected timesteps. Middle: corresponding token uncertainty maps. Right: temporally aggregated pruning statistics, including the mean map, standard deviation map, and final shared keep mask. The keep mask is computed from the aggregated uncertainty score and shared across all timesteps. Warmer colors indicate higher uncertainty, and white regions denote preserved tokens.}
	\label{fig:prune}
\end{figure}

\begin{table}[t]
\centering
\caption{Comparison of different token merge variations at two keep ratios on {CIFAR-100}.}
\label{tab:score_ablation_short}
\small
\definecolor{mygreen}{RGB}{0,150,70}
\begin{tabular}{l|l|cccc}
\toprule
\multicolumn{2}{c|}{\textbf{Method}} & \textbf{Keep ratio} & \textbf{T} & \textbf{Acc@1 (\%)} & \textbf{Acc@5 (\%)} \\
\midrule
\multicolumn{2}{c|}{QKFormer~\cite{zhou2024qkformer}} & 1.0 & 4 & 80.98 & 95.13 \\
\midrule
\multirow{4}{*}{Stage3.0} & \multirow{2}{*}{Random}
& \cellcolor{mygreen!8}$\approx 0.80$ & \cellcolor{mygreen!8}4 & \cellcolor{mygreen!8}76.89 & \cellcolor{mygreen!8}93.43\\
& & \cellcolor{mygreen!12}$\approx 0.60$ & \cellcolor{mygreen!12}4 & \cellcolor{mygreen!12}72.42 & \cellcolor{mygreen!12}91.47 \\
\cmidrule{2-6}
& \multirow{2}{*}{Uncert}
& \cellcolor{mygreen!8}$\approx 0.80$ & \cellcolor{mygreen!8}4 & \cellcolor{mygreen!8}77.97 & \cellcolor{mygreen!8}94.01 \\
& & \cellcolor{mygreen!12}$\approx 0.60$ & \cellcolor{mygreen!12}4 & \cellcolor{mygreen!12}74.36 & \cellcolor{mygreen!12}92.34 \\
\midrule
\multirow{4}{*}{Stage3.1} & \multirow{2}{*}{Random}
& \cellcolor{mygreen!8}$\approx 0.80$ & \cellcolor{mygreen!8}4 & \cellcolor{mygreen!8}77.00 & \cellcolor{mygreen!8}{93.32} \\
& & \cellcolor{mygreen!12}$\approx 0.60$ & \cellcolor{mygreen!12}4 & \cellcolor{mygreen!12}75.58 & \cellcolor{mygreen!12}{92.30} \\
\cmidrule{2-6}
& \multirow{2}{*}{Uncert}
& \cellcolor{mygreen!8}$\approx 0.80$ & \cellcolor{mygreen!8}4 & \cellcolor{mygreen!8}77.90 & \cellcolor{mygreen!8}93.47\\
& & \cellcolor{mygreen!12}$\approx 0.60$ & \cellcolor{mygreen!12}4 & \cellcolor{mygreen!12}75.25 & \cellcolor{mygreen!12}92.26 \\
\bottomrule
\end{tabular}
\end{table}

\subsection{Ablation of Temporal Uncertainty (RQ3)}

To better understand the role of temporal uncertainty in token selection, we first visualize the uncertainty patterns of representative tokens across time steps. As shown in Figure~\ref{uncertainty_heatmap}, token uncertainty is highly non-uniform over both token index and temporal dimension. While some tokens maintain relatively stable uncertainty patterns across time, others exhibit clear temporal fluctuations, indicating substantial differences in temporal uncertainty variation. This observation suggests that token importance in spiking transformers should not be estimated solely from static response magnitude or single-step statistics. Instead, temporal uncertainty provides a more informative cue for distinguishing informative tokens from redundant ones.

Based on this observation, Table~\ref{tab:temporal_short} further compares several temporal scoring strategies on CIFAR-100. Using only the average uncertainty across time already yields competitive performance, indicating that the overall uncertainty level is a useful cue for token selection. However, relying on the temporal standard deviation alone leads to noticeably inferior results, especially at a lower keep ratio, suggesting that temporal fluctuation by itself is insufficient to characterize token importance. In contrast, combining the mean and standard deviation achieves the best overall performance at both keep ratios. In particular, the proposed score reaches 81.06\% Acc@1 at keep ratio 0.8 and 80.55\% at keep ratio 0.6, outperforming the other temporal variants.

These results indicate that token uncertainty in spiking transformers is inherently temporal, and that both its overall level and temporal fluctuation carry useful information for token selection. Jointly modeling the uncertainty level and its temporal fluctuation provides a more effective criterion for selecting informative and temporally discriminative tokens, which in turn leads to better pruning performance.

\begin{figure}[t]
	\centering
\subfloat[{{Stage 3.0.}}]{
    \includegraphics[width = 0.47\linewidth]{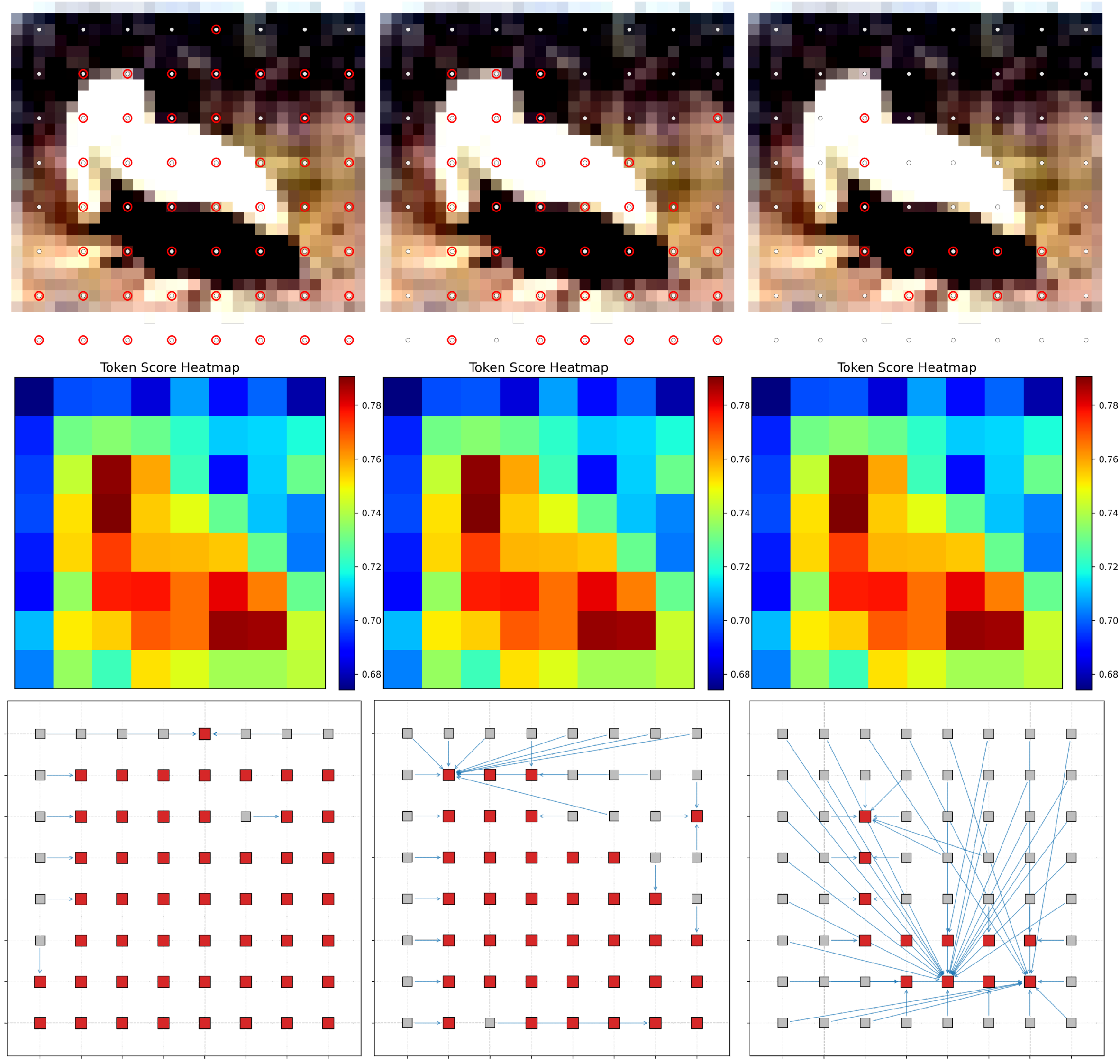}
    \label{fig:sub_3.0}
  } \hfill
  \subfloat[{Stage 3.1.}]{
    \includegraphics[width = 0.47\linewidth]{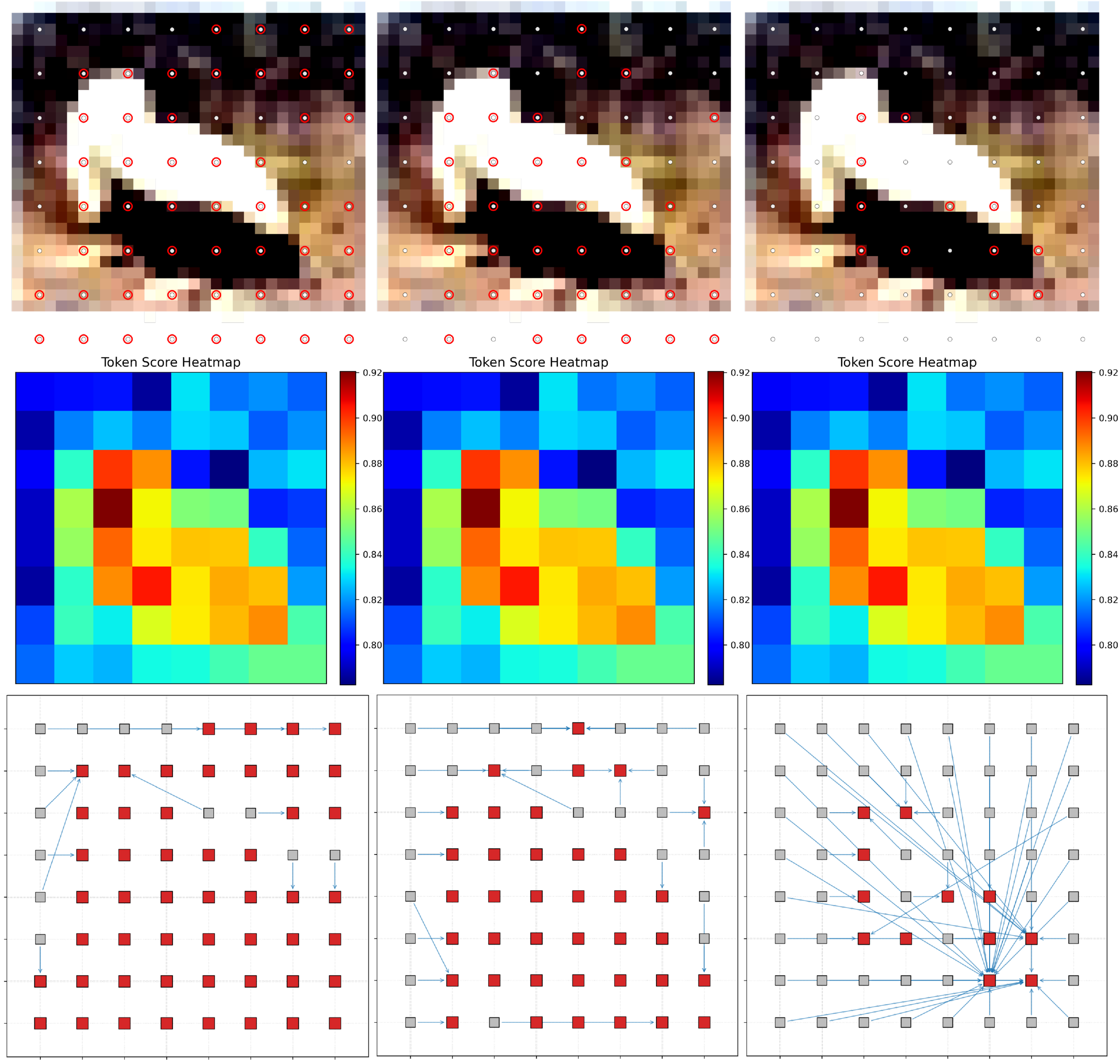} 
    \label{fig:sub_3.1}  } 
    \caption{\textbf{Merge visualization of Uncert (QKFormer) on CIFAR-100.} Top: input image with token centers, where preserved tokens are highlighted. Middle: temporally aggregated token score heatmap used for merge decisions, with warmer colors indicating higher scores. Bottom: token merge correspondence map, where red squares denote preserved tokens, gray squares denote merged tokens, and blue arrows indicate merge directions toward their assigned anchors.}
	\label{fig:heatmap}
\end{figure}

\textbf{Analysis of Prune.} Figure~\ref{fig:prune} visualizes the uncertainty maps at different timesteps and the resulting shared keep masks under two pruning ratios on DVS-CIFAR-10. The per-timestep uncertainty maps exhibit noticeable variation across time, indicating that token importance in spiking representations is inherently dynamic rather than static. The mean uncertainty maps capture tokens that remain consistently informative across timesteps, while the standard-deviation maps highlight tokens with pronounced temporal variation. By combining these statistics, the proposed method derives a shared keep mask that is not determined by any individual timestep, but by temporally aggregated uncertainty over the full sequence. As the pruning ratio becomes more aggressive, the keep mask becomes increasingly sparse, retaining only a compact set of temporally informative and discriminative tokens. This visualization supports the effectiveness of temporal uncertainty as a principled criterion for training-free token pruning.

\subsection{Ablation of Token Merge}

Table~\ref{tab:score_ablation_short} compares different token merge variants on CIFAR-100. Here, random denotes randomly selecting keep tokens as merge anchors, while the remaining tokens are merged into these anchors based on feature similarity. Overall, Uncert outperforms Random in most settings, showing that temporal uncertainty provides a useful signal for anchor selection. On stage3.0, Uncert improves Acc@1 from 76.89\% to 77.97\% at keep ratio $\approx$0.80 , and from 72.42\% to 74.36\% at $\approx$ 0.60. On stage3.1, the gain is smaller and disappears at stronger compression. Comparing insertion stages, stage3.1 is generally more suitable for merging, especially at lower keep ratios. However, all merge variants remain clearly below the baseline accuracy of 80.98\%, indicating that the proposed uncertainty is more effective for pruning than for merging.

Figure~\ref{fig:heatmap} visualizes uncertainty-guided token merging at stage3.0 and stage3.1 under three merge ratios (0.8, 0.6, and 0.2). As the merge ratio decreases, the number of preserved anchors is progressively reduced, and more tokens are assigned to a smaller set of anchors, leading to an increasingly pronounced many-to-one merging pattern. The score heatmaps show that high-score tokens are mainly concentrated around the object region rather than the background, indicating that the proposed uncertainty criterion identifies semantically informative locations. Correspondingly, the merge maps reveal that non-anchor tokens are not merged arbitrarily, but are consistently aggregated toward these informative anchors. Compared with stage3.0, stage3.1 exhibits a more compact score distribution and clearer anchor-centered aggregation, suggesting that deeper-layer tokens are more suitable for merging. However, when the merge ratio is reduced to 0.2, a large number of tokens are forced to share only a few anchors, which likely introduces information mixing and helps explain the substantial performance degradation under aggressive merging.

\subsection{Generalization on Other Baseline (RQ4)}

\begin{figure}[t]
\centering
\setlength{\tabcolsep}{2pt}
\renewcommand{\arraystretch}{1.0}
\begin{tabular}{cc}
    \includegraphics[width=0.48\linewidth]{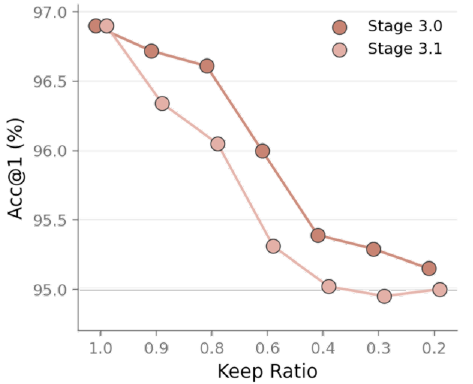} &
    \includegraphics[width=0.48\linewidth]{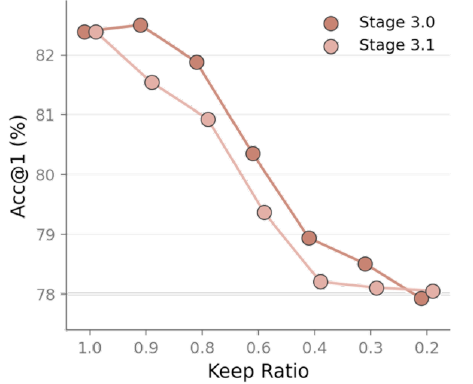} \\
    \scriptsize{(a) Accuracy \textit{vs.} keep ratio on~{CIFAR-10}.} &
    \scriptsize{(b) Accuracy \textit{vs.} keep ratio on~{CIFAR-100}.} \\[3pt]

    \includegraphics[width=0.48\linewidth]{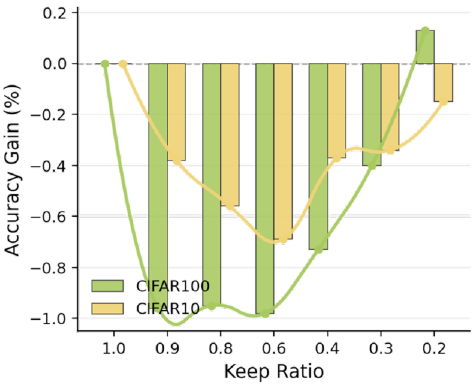} &
    \includegraphics[width=0.48\linewidth]{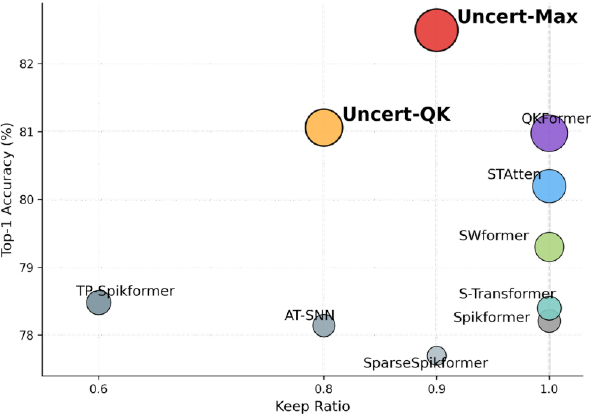} \\

    \scriptsize{(c) Accuracy gain under varying keep ratios.} &
    \scriptsize{(d) Performance comparison of different models.} \\
\end{tabular}
\caption{\textbf{Comparison across stages and methods.} Here, stage denotes the $N$-th block where token selection is applied within spiking self-attention. (a) and (b) show the Acc@1 versus keep-ratio curves when token pruning is applied at stage 3.0 and stage 3.1 on CIFAR-10 and CIFAR-100, respectively. (c) summarizes the relative accuracy changes under different keep ratios, highlighting the robustness of our method across compression levels. (d) compares our method with MaxFormer~\cite{nips2025} and other baselines.}
\label{fig:maxformer}
\end{figure}

Figure~\ref{fig:maxformer} provides a stage-wise and trade-off analysis of the proposed uncertainty-guided pruning strategy. From (a) and (b), the accuracy decreases gradually as the keep ratio becomes smaller, which is expected since more tokens are removed under stronger compression. Nevertheless, the degradation remains moderate over a broad range of keep ratios, indicating that the proposed uncertainty criterion can effectively preserve informative tokens while discarding less useful ones. Comparing stage 3.0 and stage 3.1, we observe that the two stages exhibit different sensitivity to pruning, suggesting that token redundancy and semantic compactness vary across network depth. Figure~\ref{fig:maxformer}(c) further shows that under moderate compression, the proposed method maintains stable performance and may even yield slight gains in some settings, implying that removing redundant or noisy tokens can improve representation quality. Finally, the trade-off plot in (d) shows that our method achieves a favorable balance between token retention and classification accuracy compared with representative alternatives. These results confirm that temporal uncertainty provides an effective and robust criterion for training-free token pruning in spiking transformers.

\subsection{Efficiency Analysis}
\begin{table}[t]
\centering
\caption{Comparison under different keep ratios of block-level actual SOP on our Uncert pruned stage3.1.}
\label{tab:block_sop}
\definecolor{mygreen}{RGB}{0,150,70}
\begin{tabular}{c|ccc}
\toprule
Keep ratio & Actual SOP (G) & Reduction (\%) & Power (mJ) \\
\midrule
1.00 & 1.233 & - & 1.457 \\
\cellcolor{mygreen!8}0.80 & \cellcolor{mygreen!12}1.166 & \cellcolor{mygreen!8}5.48  & \cellcolor{mygreen!12}1.441 \\
\cellcolor{mygreen!12}0.60 & \cellcolor{mygreen!12}1.104 & \cellcolor{mygreen!12}10.43 & \cellcolor{mygreen!12}1.424 \\
\cellcolor{mygreen!16}0.40 & \cellcolor{mygreen!16}1.050 & \cellcolor{mygreen!16}14.86 & \cellcolor{mygreen!16}1.406 \\
\bottomrule
\end{tabular}
\end{table}

Table~\ref{tab:block_sop} shows that the actual block-level SOP on stage3.1 decreases monotonically as the keep ratio becomes smaller. Specifically, the SOP is reduced from 1.233 G at keep=1.00 to 1.050 G at keep=0.40, corresponding to a 14.86\% reduction. This result confirms that our uncertainty-guided pruning brings measurable computation savings on the target block. Together with the accuracy results, it demonstrates a favorable trade-off between efficiency and performance under moderate pruning ratios. The total theoretical power is computed by aggregating the monitored operations of Convs, and Linear layers, along with the additional matrix multiplications and element-wise multiplications in self-attention, under a unified assumption of 0.9 $pJ$ per operation.

\section{Conclusion}
We propose Uncert, a training-free uncertainty-aware token importance estimation framework for spiking transformers. By modeling token-wise class evidence with a Dirichlet distribution and aggregating temporal uncertainty across spiking steps, Uncert provides an effective criterion for inference-time token reduction beyond single-step response cues. Experiments on static and neuromorphic benchmarks demonstrate favorable accuracy-efficiency trade-offs, with the most consistent gains under token pruning. These results suggest that temporal uncertainty offers a useful perspective for understanding token dynamics in spiking transformers and improving efficiency in a plug-and-play manner.

\noindent\textbf{Limitations.}
Although Uncert is training-free and broadly applicable, the usefulness of temporal uncertainty may still vary across architectures, layers, and pruning ratios. In particular, the optimal uncertainty aggregation and insertion stage can depend on the redundancy pattern of the backbone and the input characteristics. Future work will explore more adaptive uncertainty modeling and token reduction strategies for different spiking transformer designs.

\begin{acks}
To Robert, for the bagels and explaining CMYK and color spaces.
\end{acks}

\bibliographystyle{ACM-Reference-Format}
\bibliography{arxiv}

\end{document}